\documentclass[10pt]{article} %
\usepackage[preprint]{tmlr}

\usepackage{amsmath,amsfonts,bm}

\def\eqref#1{equation~\ref{#1}}

\def\1{\bm{1}}

\DeclareMathAlphabet{\mathsfit}{\encodingdefault}{\sfdefault}{m}{sl}
\SetMathAlphabet{\mathsfit}{bold}{\encodingdefault}{\sfdefault}{bx}{n}

\usepackage[utf8]{inputenc} %
\usepackage[T1]{fontenc}    %
\usepackage{url}            %
\usepackage{booktabs}       %
\usepackage{amsfonts}       %
\usepackage{nicefrac}       %
\usepackage{microtype}      %
\usepackage{xcolor}         %

\usepackage{xspace}

\usepackage{graphicx}
\usepackage{amsmath}
\usepackage{amssymb}
\usepackage{booktabs}
\usepackage{pifont}%
\usepackage{arydshln}

\usepackage{stfloats}
\fnbelowfloat

\usepackage{colortbl} 
\usepackage{multirow}

\usepackage{caption}
\usepackage{wrapfig}

\usepackage{tikz}
\usetikzlibrary{bayesnet}
\usetikzlibrary{arrows}
\usepackage{float}
\usepackage[all]{tcolorbox}
\usepackage{tabularx}
\usepackage{array}

\usepackage{xspace}
\newcommand{\eg}{e.g.,\@\xspace}

\usepackage{hyperref}
\usepackage{url}

\title{$\alpha$-Graph: Attention-Infused Normalizing Flow Approach to Tractable Graph Modeling}

\author{\name Thanh-Dat Truong \email tt032@uark.edu \\
      \addr CVIU Lab, Dep.  of EECS\\
      University of Arkansas
      \AND
      \name Sarah Alharbi \email sarahalh@andrew.cmu.edu \\
      \addr Dep. of Electrical and Computer Engineering\\
      Carnegie Mellon University
      \AND
      \name Susan Gauch \email sgauch@uark.edu \\
      \addr Dep.  of EECS\\
      University of Arkansas
      \AND
      \name Xinghui Zhao \email x.zhao@wsu.edu \\
      \addr Dep.  of Computer Science \\
      Washington State University Vancouver
      \AND
      \name Marios Savvides \email marioss@andrew.cmu.edu \\
      \addr Dep. of Electrical and Computer Engineering\\
      Carnegie Mellon University
      \AND
      \name Khoa Luu \email khoaluu@uark.edu \\
      \addr CVIU Lab, Dep. of EECS\\
      University of Arkansas
      }

\def\month{MM}  %
\def\year{YYYY} %
\def\openreview{\url{https://openreview.net/forum?id=XXXX}} %

\begin{document}

\maketitle

\begin{abstract}
Graph modeling, a crucial task for representing complex relationships in graph-structured data, has achieved significant success in recent years. However, current graph modeling methods rely on traditional Graph Neural Networks and pre-training approaches to implicitly learn the underlying relational structure of graph data. Thus, these prior methods cannot capture the complex graph structure and correlations among inputs. In this paper, we introduce a novel Attention-based Normalizing Flow-based Approach\footnote{Our implementation and models will be released publicly for research reproducibility.} (ANFA or $\alpha$) that provides an explicit, interpretable, and tractable Graph Modeling ($\alpha$-Graph). In particular, we propose a new Unconditional Graph Normalizing Flow with an Invertible Attention Mechanism to capture the complex relational structure of graph data. To further enhance the expressiveness of the model, we introduce Conditional Graph Normalizing Flow with Learnable Queries that enables efficient modeling of correlations in graph-structured data. We show that our Conditional Graph Normalizing Flows behave similarly to Unconditional Graph Normalizing Flows, enhancing expressiveness while maintaining training stability and efficiency. Our experimental results on three benchmarks will illustrate the effectiveness and the state-of-the-art (SoTA) performance of the proposed $\alpha$-Graph method.
\end{abstract}

\section{Introduction}

\begin{figure}[!b]
    \centering
    \vspace{-2mm}
    \includegraphics[width=1.0\linewidth]{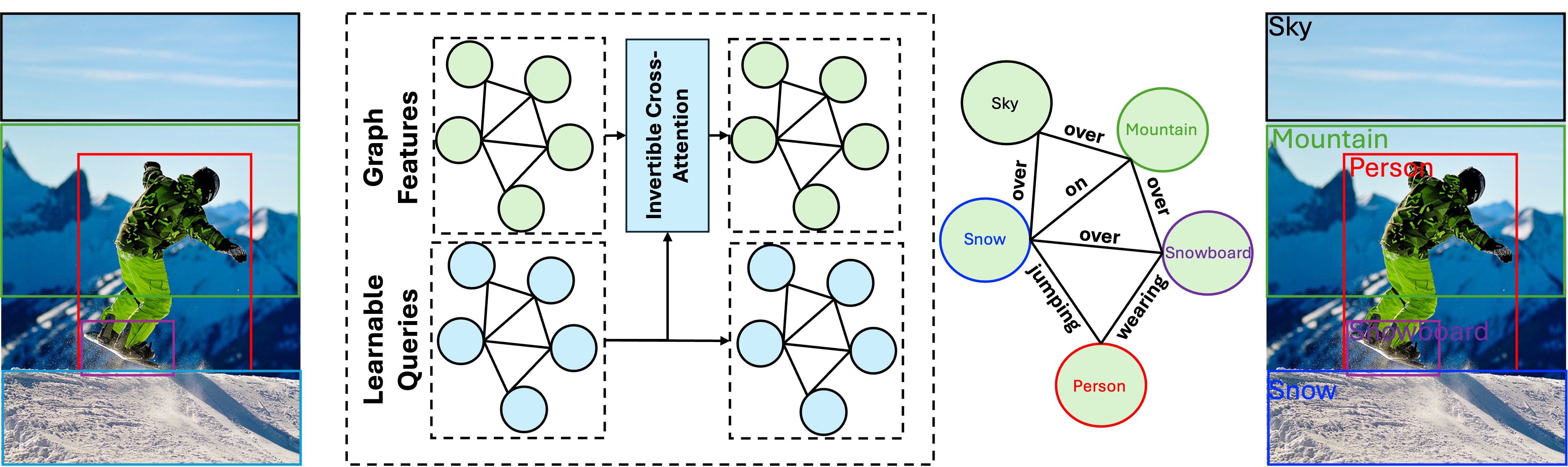}
    \vspace{-4mm}
    \caption{\textbf{Our Conditional Normalizing Flow-based Approach Via  Invertible Cross-Attention with Learnable Queries.}}\label{fig:highlight}
\end{figure}

Various practical problems, \eg Scene Graph Generation \cite{yang2022panoptic, nguyen2024cyclo, nguyen2024hig}, Action Recognition \cite{shi2019skeleton, shi2019two}, 
can be represented as graphs that model entities and their complex interrelationships.
Modeling these dependencies in graph-structured data is important because relationships among nodes often encode critical feature representations that directly influence the outcomes of prediction and reasoning tasks.
Graph neural networks (GNNs) \cite{kipf2016semi, kipf2016variational, kim2024hypeboy} have emerged as a powerful approach to graph modeling, yielding performance improvements by integrating graph topology into the learning process. 
This approach enables various applications, \eg relationship modeling \cite{nguyen2024cyclo, nguyen2024hig}, molecular and drug discovery \cite{ruan2021exploring, saifuddin2023hygnn}, and robotics \cite{pistilli2023graph, lei2023graph}.

The success of graph modeling relies on advances in pre-training strategies and network design. Indeed, pre-training \cite{kipf2016variational, hou2022graphmae, hou2023graphmae2, kim2024hypeboy} enables GNNs to model generalizable structural and relational features from large-scale, unlabeled graphs, providing a strong initialization for downstream applications. Meanwhile, recent studies have introduced various network architectures for graphs, e.g., Graph CNN \cite{kipf2016semi, huang2021unignn}, Graph Transformer \cite{yun2019graph, velivckovic2017graph}, to enhance GNNs' ability to capture complex dependencies and hierarchical relationships in graph-structured data. 
However, these prior approaches \cite{kipf2016variational, hou2022graphmae, hou2023graphmae2, kim2024hypeboy} primarily rely on implicit modeling, where representations are learned directly from data without explicit encoding of domain knowledge or structural constraints. 
In particular, these models often lack interpretability, making it difficult to understand which structural features they capture. 
Graph Isomorphism Network \cite{xu2018powerful} or Graph Attention Network \cite{velivckovic2017graph} rely purely on node aggregation without semantic decomposition. 
Deeper GNNs may also encounter the oversmoothing problem, where node representations become indistinguishable, and performance degrades, especially in large or sparse graphs \cite{rusch2023survey}.
Other implicit pre-training methods, \eg GraphVAE \cite{kipf2016variational, zhang2019d} and contrastive learning \cite{xia2022simgrace, ju2024towards}, may not align well with downstream goals since these learning objectives aim to reconstruct graph structures or maximize invariance under augmentations, rather than learning task-specific.
Moreover, implicit modeling often relies on learning from large-scale data without enforcing structural priors \cite{kim2024hypeboy, hou2022graphmae}. As a result, the model can overfit to dataset-specific patterns, rather than modeling generalizable relationship structures~\cite{rusch2023survey}.

While most recent graph modeling methods adopt attention mechanisms or message-passing schemes to implicitly capture structural dependencies, explicit modeling approaches have gained less attention.
Normalizing Flows \cite{dinh2016density, kingma2018glow, sukthanker2022generative, truong2025mango} offer a promising approach for explicit modeling by learning exact likelihoods through invertible mappings between graph-structured data and latent spaces. By modeling the graphs via a bijection, normalizing flows enable stable and tractable training while capturing complex relational structures.
In addition, this approach can represent complex graph distributions, capturing relational features and graph topologies, with direct control over model parameters \cite{kingma2018glow, sukthanker2022generative}.
This explicit approach also enhances interpretability and improves understanding of graph-structured dependencies, a limitation of prior implicit models \cite{dinh2016density, ho2019flow++}. 
Compared to traditional methods \cite{hou2023graphmae2, hou2022graphmae, kipf2016variational, zhang2019d}, explicit modeling with normalizing flows offers a more effective mechanism for capturing graph structures without overemphasizing node-specific features. 
Consequently, it enables more precise, flexible, and robust graph modeling, improving generalization.

\noindent
\textbf{Limitations of Normalizing Flows in Graph Modeling.}
The properties of graph-structured data, e.g., permutation invariance, discrete representation, or relational dependencies, pose several challenges in designing an invertible network.
Meanwhile, normalizing flows were initially designed for continuous or image spaces \cite{dinh2014nice, dinh2016density}.
Current normalizing flow designs rely on the coupling layer \cite{dinh2016density}.
However, it remains limited for graph-structured data.
Indeed, the scale and transformation networks \cite{dinh2016density} in coupling layers are designed using residual networks \cite{he2016deep}, which cannot capture the relational structure of the graph.
While GNNs \cite{kipf2016semi} can capture relational structure, these layers are not invertible.
Recent work \cite{liu2019gnf} introduced invertible layers for GNNs based on coupling layers, the expressiveness of these layers is limited due to the fixed feature-dimension partition mechanism \cite{kingma2018glow}. 
Other studies have adopted autoregressive modeling with RNNs \cite{li2018learning, you2018graphrnn} to model the likelihood of graph data, but this remains intractable and cannot capture long-range dependencies.
Therefore, to develop an efficient normalizing flow approach to graph modeling, it is critical to address these limitations.

\noindent
\textbf{Contributions of this Work.}
This paper introduces the novel \textit{Attention-based Normalizing Flow-based Approach} \textbf{($\alpha$-Graph)} to develop a new explicit, interpretable, and tractable Graph Modeling (Figure \ref{fig:highlight}).
Our contributions can be summarized as follows.
First, we propose a new Invertible Attention layer for Graph Normalizing Flow Models. 
Our proposed invertible attention layer efficiently addresses the limitations of prior coupling layers while maintaining traceability and invertibility and enforcing graph-structured properties.
Second, to further enhance the modeling capability of the graph normalizing flow, we propose a Conditional Normalizing Flow for Graph Modeling by introducing learnable queries and invertible cross-attention layers. Our conditional graph normalizing flow model can efficiently capture the correlations in graph-structured data. 
Third, we prove that our Conditional Graph Normalizing Flows have a gradient update similar to that of Unconditional Graph Normalizing Flows.
This demonstrates that $\alpha$-Graph enhances expressiveness while inheriting the strong optimization properties of unconditional graph normalizing flows, with theoretical guarantees.
Finally, extensive experiments on diverse graph datasets demonstrate the effectiveness of $\alpha$-Graph across various aspects, achieving State-of-the-Art (SoTA) performance compared to prior methods in graph modeling.

\section{Related Work and Background}
\label{sec:related_work}

\vspace{-1mm}
\subsection{Related Work}
\vspace{-1mm}

\noindent
\textbf{Graph Learning.} 
GNNs generate informative graph embeddings by jointly leveraging node features and structural information. Common GNNs, (e.g., GCN \cite{kipf2016semi}, GAT \cite{velivckovic2017graph, yun2019graph}, GIN \cite{xu2018powerful}, or GraphSAGE \cite{hamilton2017inductive}, TopoFormer \cite{uddin2026topoformer}, ReIGT \cite{dwivedi2026relational}) follow a message-passing paradigm, where each node updates its representation by aggregating information from its neighbors. 
Later, to further improve the modeling capability of GNNs, several self-supervised learning approaches have been introduced.
Several methods introduced the auto-encoder framework the graph representation learning, e.g., VGAE \cite{kipf2016variational}, ARVGA \cite{pan2018adversarially}, MGAE \cite{tan2022mgae}, GALA \cite{park2019symmetric, chen2023does}, GATE \cite{salehi2019graph}, or NWR-GAE \cite{tang2022graph}.
GraphMAE and GraphMAE2 \cite{hou2022graphmae, hou2023graphmae2} presented a masked feature reconstruction task during the pre-training phase.
Other methods adopted autoregressive modeling (e.g., GraphRNN \cite{you2018graphrnn}, GCPN \cite{you2018graph}, and GPT-GNN \cite{hu2020gpt}) or diffusion (e.g., LR-FGDM \cite{wang2026lowrank}, GGND \cite{hong2026ggnd}, QAFD-RAG \cite{zhou2026queryaware}, DualDiff \cite{xie2026global}, HOG-Diff \cite{huang2026hogdiff}) to learn representations in graph-structure data.
Recent studies \cite{kim2024hypeboy, huang2021unignn, huang2026hyper} used hypergraph modeling to further improve performance by exploring higher-order graph structures.

\noindent
\textbf{Attention Models.} 
The attention mechanism in Transformers has proven highly effective in various tasks \cite{vaswani2017attention} because it can model long-range dependencies via second-order correlations. Transformers typically employ two types of attention: self-attention, which captures intra-relationships \cite{vaswani2017attention}, and cross-attention, which models inter-interactions. This flexibility has made Transformers a preferred design in various problems \cite{dosovitskiy2020image, devlin2018bert, velivckovic2017graph, bjorck2025gr00t, nguyen2024insect, truong2022direcformer, truong2021right2talk, truong2023fredom, truong2025cross, truong2025falcon, truong2025insect}.

\noindent
\textbf{Explicit Modeling via Normalizing Flows.}
RealNVP \cite{dinh2016density} first introduced the affine coupling layer that enables efficient computation of both the inverse transformation and the log-determinant of the Jacobian. 
Later, NICE \cite{dinh2014nice} introduced non-linear independent component estimation. Meanwhile, Glow \cite{kingma2018glow} incorporated invertible $1 \times 1$ convolutions, activation normalization, and multi-scale architectures to enhance performance and stability. 
Additionally, autoregressive flows \cite{huang2018neural} and equivariant flows \cite{garcia2021n} have been explored to improve modeling capacity.
To further increase expressiveness, recent approaches have integrated attention-based mechanisms as scale and translation networks \cite{ho2019flow++, sukthanker2022generative}. 
Prior studies also adopted normalizing flows for conditional modeling \cite{lu2020structured, lugmayr2020srflow}, either using a single conditional invertible network \cite{sorkhei2021full} or deploying dual invertible architectures \cite{sun2019dual} to model joint distributions.
GRevNets \cite{liu2019gnf} presented normalizing flows for graph modeling by using GNNs in coupling layers.
While these models improve the ability to model complex distributions, they still struggle with capturing long-range dependencies and relational structure.

\vspace{-1mm}
\subsection{Normalizing Flow and Its Limitation in Graph Modeling}\label{ref:limitation-nvp}
\vspace{-1mm}

Given a graph $\mathcal{G} = (\mathcal{V}, \mathcal{E})$ where $\mathcal{V} = \{v_i\}_{i=1}^N$ is the set of nodes, $\mathbf{E}$ is the set of edges defining the relational structure of the graph, $N = |\mathcal{V}|$ is the number of nodes, we define $\mathbf{X} = [\mathbf{x}_1, \mathbf{x}_2, ..., \mathbf{x}_N]$ are the features of nodes, i.e., $\mathbf{x}_i$ is features of vertex $v_i$. 
The normalizing flow-based model \cite{dinh2016density, dinh2014nice, kingma2018glow} for graph-structured data can be designed via the invertible affine couple layer as follows,
\begin{equation}
\begin{split}
    \mathbf{X}_1, \mathbf{X}_2 &= \operatorname{split}(\mathbf{X}), \quad \mathbf{Y}_1 = \mathbf{X}_1, \quad \mathbf{Y}_2 = \mathbf{X}_2 \odot \exp\left(\mathcal{S}(\mathbf{X}_1)\right) + \mathcal{T}(\mathbf{X}_1) \quad \Rightarrow \mathbf{Y} = \operatorname{join}([\mathbf{Y}_1, \mathbf{Y}_2])
\end{split}
\end{equation}
$\operatorname{split}$ is a splitting approach to divide the inputs into two parts, e.g., feature-dimension splitting. $\mathcal{S}$ and $\mathcal{T}$ are deep neural networks for scaling and translation, $\operatorname{join}$ is a joining method, and $\odot$ is the element-wise matrix multiplication.

\noindent
\textbf{Limitations.} The expressiveness of the flow-based network in graph modeling relies on the design of the invertible layers. However, the current design of the coupling layers \cite{dinh2016density, dinh2014nice, kingma2018glow, sukthanker2022generative, ho2019flow++} is inefficient in modeling graph-structured data. In particular, the common approaches \cite{dinh2014nice, dinh2016density, truong2021bimal, nhan2017temporal, duong2020vec2face} adopted residual networks for the scaling and translation network. Still, this network design cannot capture the relational structure of the graph defined by the list of edges $\mathcal{E}$.
While recent work adopted a graph neural network for $\mathcal{S}$ and $\mathcal{T}$, the splitting mechanism of $\operatorname{split}$ based on the feature-dimension splitting limits the expressiveness and flexibility of the model by only allowing partial updates of the input at each layer.
In addition, these prior methods \cite{dinh2016density, kingma2018glow} cannot capture long-range dependencies of the graph-structured data. Therefore, this work develops a novel Attention-based Normalizing Flow-based Graph Modeling approach to address these limitations.

\section{The Proposed $\alpha$-Graph Approach}

In this section, we first model graphs with unconditional graph normalizing flows via Invertible Attention to encode graph structure while preserving invertibility. We then extend to a Conditional Graph Normalizing Flow, showing it retains the same optimal solutions as the unconditional objective.

\subsection{Unconditional Graph Normalizing Flow}

\begin{figure}[!b] %
    \centering
    \vspace{-4mm}
    \includegraphics[width=1.0\linewidth]{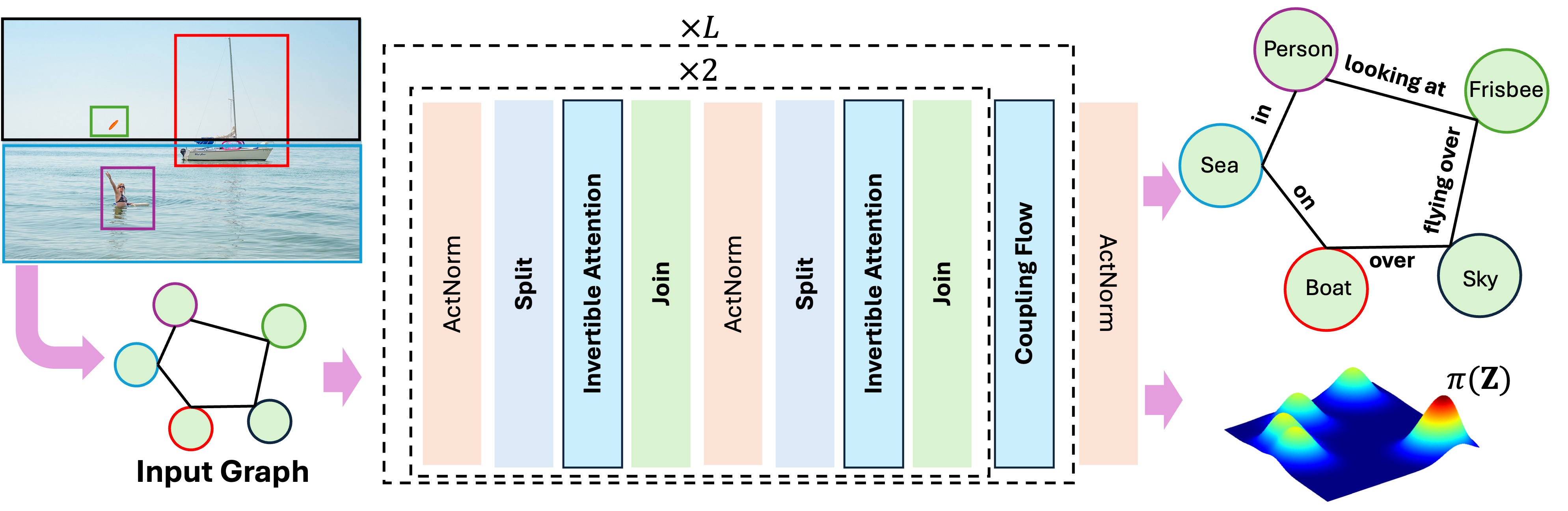}
    \vspace{-5mm}
    \caption{\textbf{Our $\alpha$-Graph Learning Framework.}}
    \label{fig:general-framework}
\end{figure}
Most recent graph modeling methods adopt Graph Neural Networks (GNNs) as the backbone for pre-training and downstream tasks \cite{hou2022graphmae, hou2023graphmae2}.
While GNNs have shown strong capabilities in learning node representations through message passing \cite{kipf2016semi}, prior studies suggest that these approaches remain limited in modeling complex graph structures and higher-order dependencies \cite{dwivedi2023benchmarking}. 
In particular, GNN-based methods primarily rely on local aggregation, which struggles to capture global relationships across nodes and the relational structure of graphs, especially when the edges $\mathcal{E}$ in the graph are sparse or noisy \cite{dwivedi2023benchmarking}.
In addition, because graphs often exhibit complex structural distributions, designing models that effectively capture diverse patterns in complex graphs remains a major challenge.
To address these limitations, we propose to \textit{model the graph structures through explicit distributions}.
In particular, we adopt Normalizing Flow Models, which provide a flexible, exact likelihood-based estimation of complex distributions via invertible transformations.
By learning a bijective mapping between latent spaces and the graph space, our framework can effectively capture the complex dependencies in graph-structured data.

Formally, let $G$ be the bijective network that maps the input features $\mathbf{X}$ of graph $\mathcal{G}$ into the latent space, i.e., $\mathbf{Z} = G(\mathbf{X})$.
Figure \ref{fig:general-framework} illustrates an overview of our proposed Normalizing Flow-based framework for Graph Modeling.
Then, the graph data distribution $p(\mathbf{X})$ can be formed via the Normalizing Flow-based Model $G$ as $\log p(\mathbf{X}) = \log \pi(\mathbf{Z}) + \log \left|
    \frac{\partial G(\mathbf{X})}{\partial \mathbf{X}}\right|$, 
where $\pi(\mathbf{Z})$ is the prior Normal distribution.
For the downstream tasks, the prediction $\mathbf{\hat{Y}}$ is generated by applying a projection head to the latent representation $\mathbf{Z}$, i.e., $\mathbf{\hat{Y}} = \operatorname{Head}(\mathbf{Z})$, where $\operatorname{Head}$ denotes a task-specific module (e.g., classification head) that maps latent features to the corresponding outputs.
The projection $\operatorname{Head}$ can be learned by fine-tuning the model $G$ after pre-training.
The success of graph normalizing flows relies on the design of $G$. In the next section, we introduce new Invertible Attention layers to address the limitation of prior coupling layers.

\subsection{Invertible Attention in Graph Modeling}

\begin{wrapfigure}[12]{r}{0.48\linewidth}
    \centering
    \vspace{-4mm}
    \includegraphics[width=1.0\linewidth]{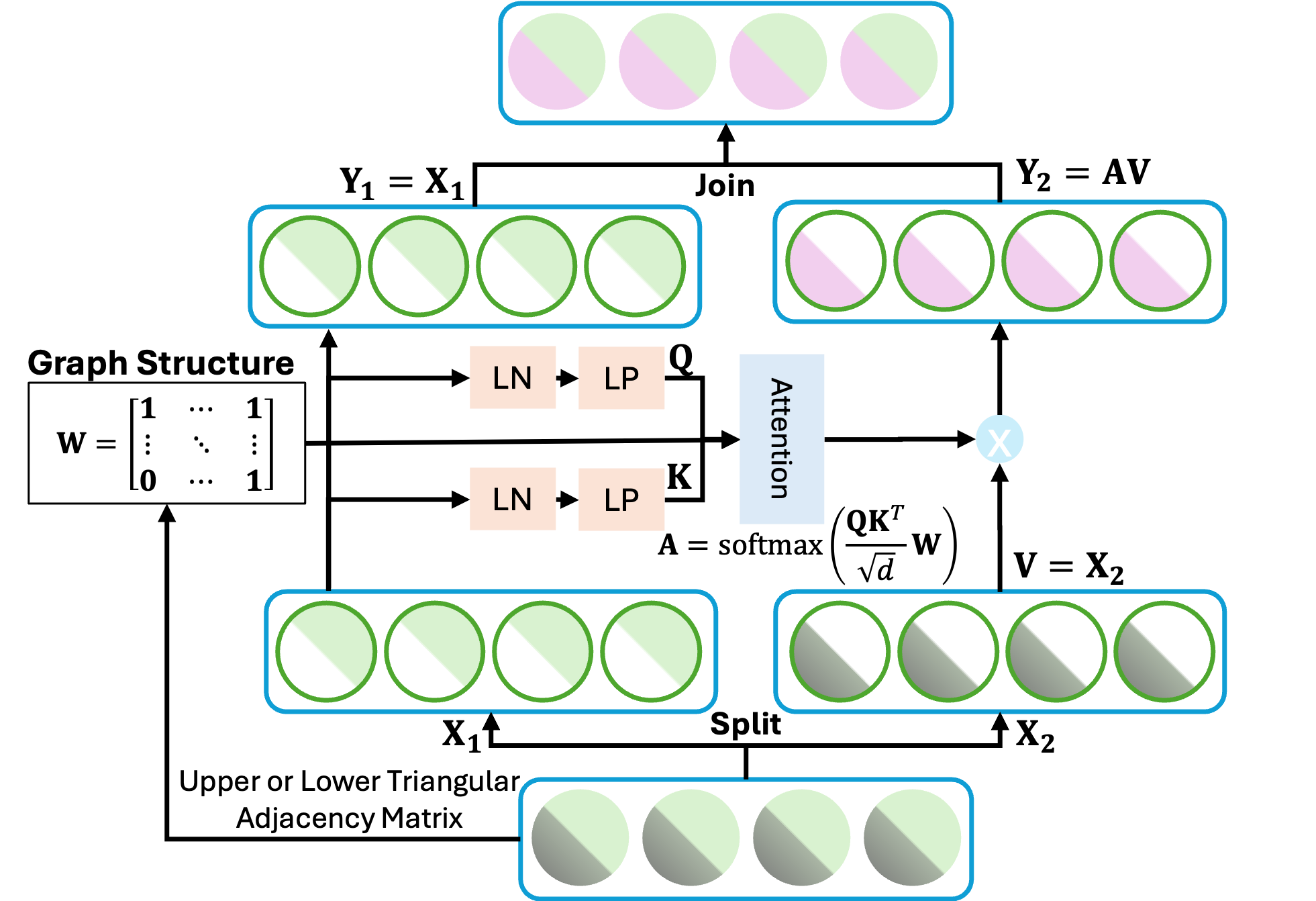}
    \vspace{-6mm}
    \caption{Our Invertible Attention Mechanism.}
    \label{fig:attention-mechanism}
    \vspace{-4mm}
\end{wrapfigure}
We propose a novel Invertible Attention mechanism to address the limitations of coupling layers in normalizing flow-based models presented in Sec. \ref{ref:limitation-nvp}. The effectiveness of attention mechanisms fundamentally depends on their ability to capture and model relationships among feature representations through second-order correlations. 
Formally, the attention layers can be formally defined as Eqn.\eqref{eqn:attention}.
\begin{equation}
\label{eqn:attention}
\begin{split}
    \mathbf{Q} &= \operatorname{LP}_{Q}(\operatorname{LN}(\mathbf{X})) \; \mathbf{K} = \operatorname{LP}_{K}(\operatorname{LN}(\mathbf{X})) \; \mathbf{V} = \operatorname{LP}_{V}(\operatorname{LN}(\mathbf{X})) \\
    \mathbf{Y} &= \operatorname{Attention}(\mathbf{Q}, \mathbf{K}, \mathbf{V}) = \operatorname{softmax}\left(\frac{\mathbf{Q}\times\mathbf{K}^T}{\sqrt{d}}\right)\mathbf{V}
\end{split}
\end{equation}
where $\mathbf{Y}$ is the output of the attention layer, 
${d}$ is the feature dimension, 
$\operatorname{LN}$ is the layer normalization, 
$\mathbf{Q}$, $\mathbf{K}$, and $\mathbf{V}$ are the query, key, and values features obtained via three linear projections $\operatorname{LP}_{Q}$, $\operatorname{LP}_{K}$, and $\operatorname{LP}_{Q}$.
The attention weights, computed via a scale-dot product between query and key features,  
measure the pairwise relationships between features, effectively modeling their second-order dependencies. 
Then, the attention weights are applied to the value features, producing the final output aggregating the learned correlations across the features.
This mechanism ensures that the resulting features are enriched with contextual information derived from the correlations among tokens.

\noindent
\textbf{The Proposed Invertible Attention.} Motivated by this attention design, we introduce the Invertible Attention mechanism (Figure \ref{fig:attention-mechanism}) within the coupled layer as in Eqn. \eqref{eqn:invert-cross-attention}.
\begin{equation}\label{eqn:invert-cross-attention}
\begin{split}
    \mathbf{X}_1, \mathbf{X}_2 &= \operatorname{split}([\mathbf{x}_1, ..., \mathbf{x}_N]), \quad
    \mathbf{Q} = \operatorname{LP}_{Q}(\operatorname{LN}(\mathbf{X}_1)), \;
    \mathbf{K} = \operatorname{LP}_{K}(\operatorname{LN}(\mathbf{X}_1)), \;
    \mathbf{V} = \mathbf{X}_2 \\
    \mathbf{Y}_1 &= \mathbf{X}_1,  \mathbf{Y}_2 = \operatorname{softmax}\left(\frac{\mathbf{Q}\times\mathbf{K}^T}{\sqrt{d}}\right)\mathbf{V} \quad \Rightarrow
    \mathbf{Y}  = \operatorname{join}([\mathbf{Y}_1, \mathbf{Y}_2])
\end{split}
\end{equation}

In our approach, ${d}$ is a learnable variable that captures a general scale \cite{sukthanker2022generative, vaswani2017attention}.
To maintain the symmetry for invertibility, our $\operatorname{join}$ splits the inputs along the feature dimension \cite{dinh2016density, dinh2014nice, kingma2018glow}.
Our proposed layer captures inter-token interactions through the attention mechanism. In particular, the attention information extracted from the first part of inputs $\mathbf{X}_1$ is embedded into the second part $\mathbf{X}_2$.
Then, by stacking multiple invertible attention layers and systematically alternating the feature partitions, our proposed approach efficiently models feature correlations across inputs.
However, \textit{three critical factors need to be addressed in this attention mechanism.}
First, the normalizing flow-based model requires that each layer be invertible.
Second, the invertible layer must be tractable, such that the log-determinant of its Jacobian matrix can be computed efficiently.
Third, the attention mechanism should be capable of modeling the relational structure via edges $\mathcal{E}$.

\noindent
\textbf{Enforcing Graph Structure and Invertibility.}
To ensure the graph structure of $\mathcal{G}$ is embedded in the invertible attention layer, let us define the matrix $\mathbf{W}$ that is the adjacency matrix of the graph $\mathcal{G}$ constructed from the list of edges $\mathcal{E}$. Without a strict argument, we assume that the graph $G$ is an undirected graph and each node $v$ in $\mathcal{G}$ is connected to itself, i.e., $W_{i, i} = 1$.
In addition, let $\mathbf{W}_{\operatorname{up}}$ and $\mathbf{W}_{\operatorname{low}}$ be the upper and lower triangular matrices of $\mathbf{W}$.
These triangular matrices encode directional structures within the graph $\mathcal{G}$, i.e., $\mathbf{W}_{\operatorname{up}}$ captures connections from nodes with lower indices to nodes with higher indices, while $\mathbf{W}_{\operatorname{low}}$models the reverse relationships, from higher-indexed nodes to lower-indexed nodes.
This decomposition provides a structured way to represent bidirectional interactions within the graph, enabling more flexible modeling of directional dependencies across different layers or data partitions.
Then, to enforce the graph structure on our attention mechanism in Eqn. \eqref{eqn:invert-cross-attention}, we apply the the attention mask  $\mathbf{W}_{\operatorname{up}}$ as $\mathbf{Y}_2 = \operatorname{softmax}\left(\frac{\mathbf{Q}\times\mathbf{K}^T}{\sqrt{d}}\mathbf{W}_{\operatorname{up}}\right)\mathbf{V}$.
Since the graph is undirected, represented by the combination of both $\mathbf{W}_{\operatorname{up}}$ and $\mathbf{W}_{\operatorname{low}}$, we apply two consecutive invertible attention layers where the first and second attention layers use $\mathbf{W}_{\operatorname{up}}$ and $\mathbf{W}_{\operatorname{low}}$, respectively. 
Under this form, the inverse function of our invertible attention can be formed as:
\begin{equation} \label{eqn:newICA}
\begin{split}
    \mathbf{Y}_1, \mathbf{Y}_2 &= \operatorname{split}([\mathbf{y}_1, ..., \mathbf{y}_N]), \quad\quad
    \mathbf{Q} = \operatorname{LP}_{Q}(\operatorname{LN}(\mathbf{Y}_1)), \quad
    \mathbf{K} = \operatorname{LP}_{K}(\operatorname{LN}(\mathbf{Y}_1)), \quad
    \mathbf{V} = \mathbf{Y}_2 \\
    \mathbf{X}_1 &= \mathbf{X}_1, \quad \mathbf{X}_2 = \left[\operatorname{softmax}\left(\frac{\mathbf{Q}\times\mathbf{K}^T}{\sqrt{d}}\mathbf{W}_{\operatorname{up}}\right)\right]^{-1}\mathbf{V} \quad \Rightarrow
    \mathbf{X} = \operatorname{join}([\mathbf{X}_1, \mathbf{X}_2])
\end{split}
\end{equation}
The inverse layer of attention using $\mathbf{W}_{\operatorname{low}}$ is performed similarly.
Let $\mathbf{A}$ be the attention matrix, i.e., .$\mathbf{A} = \operatorname{softmax}\left(\frac{\mathbf{Q}\times\mathbf{K}^T}{\sqrt{d}}\mathbf{W}_{\operatorname{up}}\right)$.
With our design, the graph structure will be embedded into the attention layer, as $\mathbf{W}_{\operatorname{up}}$ (and $\mathbf{W}_{\operatorname{low}}$) captures the relational structure of $\mathcal{G}$.
In addition, since both $\mathbf{W}_{\operatorname{up}}$ and $\mathbf{W}_{\operatorname{low}}$ are non-zero diagonal triangular matrices, the inverse matrix of $\mathbf{A}$ will always exist.
Therefore, our approach guarantees the efficient invertibility of the attention layers.

\noindent
\textbf{Tractability.} 
As aforementioned, the invertible layer of the Normalizing Flow-based model requires the tractable computation of the determinant of the Jacobian matrix, i.e., $\det\left(\frac{\partial \mathbf{Y}}{\partial \mathbf{X}}\right)$. In our proposed approach, we can explicitly and efficiently formulate the Jacobian determinant, as shown in Eqn. \eqref{eqn:detnew}.
\begin{equation} \label{eqn:detnew}
\det\left(\frac{\partial \mathbf{Y}}{\partial \mathbf{X}}\right) =  \left(\det(\mathbf{A})\right)^{N/2} = \det \left(\operatorname{softmax}\left(\frac{\mathbf{Q}\times\mathbf{K}^T}{\sqrt{d}}\mathbf{W}_{\operatorname{up}}\right)\right)^{N/2}
\end{equation}
Since $\mathbf{A}$ is either an upper or lower triangular matrix due to $\mathbf{W}_{\operatorname{up}}$ or $\mathbf{W}_{\operatorname{low}}$, the determinant reduces to the simple product of diagonal entries of the matrix, enabling efficient computation.

\noindent
\textbf{Limitations of Feature-Dimension Splitting.}
The simple splitting mechanism employed in coupling layers \cite{dinh2014nice, dinh2016density, kingma2018glow, ho2019flow++, liu2019gnf} offers significant computational benefits. It helps the model ensure that the Jacobian remains triangular, enabling efficient log-determinant computation and straightforward invertibility.
However, this splitting strategy introduces a major limitation, i.e., the split mechanism is entirely independent of the input data and lacks adaptability to its underlying structure. As a result, the model cannot capture complex dependencies across nodes, potentially limiting its expressiveness, especially in graph-structured data.

\subsection{Conditional Graph Normalizing Flow with Learnable Queries}

\textbf{Conditional Graph Normalizing Flows with Learnable Queries.}
Learnable queries have emerged as a powerful component in attention-based architectures \cite{carion2020end, devlin2018bert}. 
Inspired by this design, we propose a new conditional normalizing flow approach with learnable queries to address the limitations of splitting along the feature dimension. 
In particular, let $\mathbf{\bar{X}} = [\mathbf{\bar{x}}_1, \mathbf{\bar{x}}_1, ..., \mathbf{\bar{x}}_N]$ be the learnable queries, where $\mathbf{\bar{x}_i} \in \mathbb{R}^{d}$ is the learnable query of vertex $i$. Then, to model the graph $\mathbf{X}$, the conditional normalizing flow-based model can be formed as $\log p(\mathbf{X} | \mathbf{\bar{X}}) = \log \pi(\mathbf{Z}|\mathbf{\bar{X}}) + \log \left|
    \frac{\partial G(\mathbf{X}, \mathbf{\bar{X}})}{\partial \mathbf{X}}\right|$, 
where $\pi(\mathbf{Z}|\mathbf{\bar{X}})$ is the conditional prior. Following the common practice, we adopt the conditional Normal distribution for our prior, i.e., $\pi(\mathbf{Z}|\mathbf{\bar{X}}) = \mathcal{N}(\mathbf{Z}; \mu(\mathbf{\bar{X}}), \sigma(\mathbf{\bar{X}}))$.
Our design with learnable queries offers several benefits. 
First, it decouples queries from the graph structure, enabling the model to selectively attend to relevant features.
Second, our proposed learnable queries introduce an implicit inductive bias that facilitates faster convergence and improved performance by encouraging the model to focus on semantically meaningful graph data structures.
Third, our approach addresses the limitation of feature-dimension splitting mechanisms as discussed in Sec. \ref{ref:limitation-nvp}.

\begin{wrapfigure}[9]{r}{0.45\linewidth}
    \centering
    \vspace{-4mm}
    \includegraphics[width=1.0\linewidth]{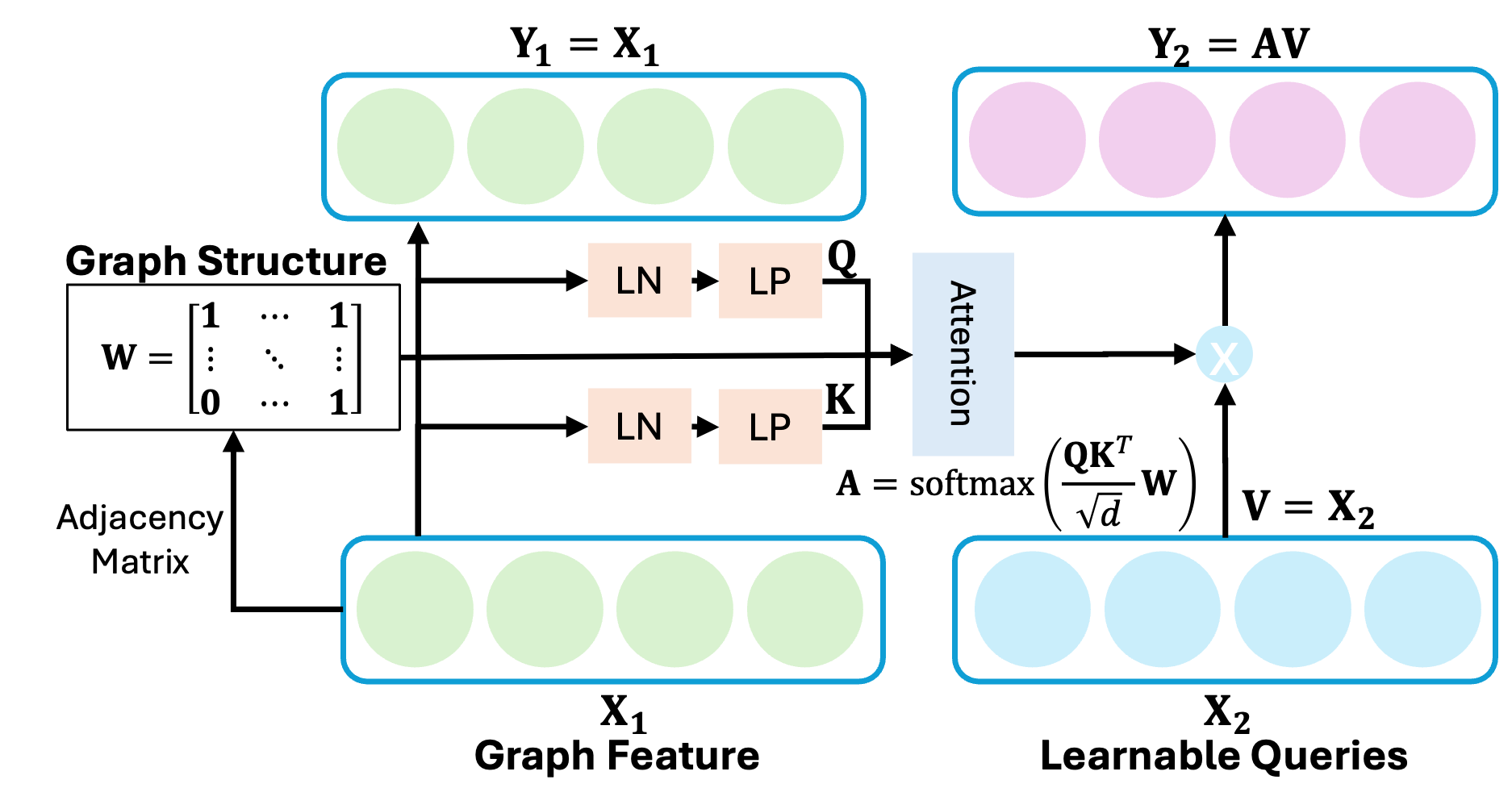}
    \vspace{-6mm}
    \caption{Our Invertible Cross-Attention Mechanism with Learnable Queries.}
    \label{fig:cross-attention-mechanism}
    \vspace{-6mm}
\end{wrapfigure}
\noindent
\textbf{Invertible Cross-Attention with Learnable Queries.} To design the invertible network $G$ with learnable queries, we propose the invertible cross-attention  (Figure \ref{fig:cross-attention-mechanism}) layer based on our invertible layer. Formally, the invertible cross-attention layer is defined as in Eqn. \eqref{eqn:att-learn-queries}.
\begin{equation}\label{eqn:att-learn-queries}
\begin{split}
    \mathbf{X}_1 &= [\mathbf{x}_1, ..., \mathbf{x}_N]  \quad \mathbf{X}_2 = [\mathbf{\bar{x}}_1, ..., \mathbf{\bar{x}}_N]  \quad \text{or vice versa} \\
    \mathbf{Q} &= \operatorname{LP}_{Q}(\operatorname{LN}(\mathbf{X}_1)), 
    \quad \mathbf{K} = \operatorname{LP}_{K}(\operatorname{LN}(\mathbf{X}_1)), \quad
    \mathbf{V} = \mathbf{X}_2 \\
    \mathbf{Y}_1 &= \mathbf{X}_1,  \quad \mathbf{Y}_2 = \operatorname{softmax}\left(\frac{\mathbf{Q}\times\mathbf{K}^T}{\sqrt{d}}\mathbf{W}_{\operatorname{up}}\right)\mathbf{V} \quad \text{(or } \mathbf{W}_{\operatorname{low}} \text{)} \\
\end{split}
\end{equation}
It should be noted that since the cross-invertible attention layer adopted the design of the invertible attention layer in Eqn. \eqref{eqn:invert-cross-attention}, the traceability and invertability of our layer in Eqn. \eqref{eqn:att-learn-queries} is still maintained. 
In our cross-attention layer, the traditional splitting mechanism is replaced by assigning the first part to the node features and the second to the learnable queries, with the roles alternating across layers. Under this design, the attention information from the node features $\mathbf{X}_1$ is captured and embedded into the learnable queries $\mathbf{X}_2$. Similarly, when the roles of $\mathbf{X}_1$ and $\mathbf{X}_2$ are interchanged, the correlations among the learnable queries are effectively modeled and propagated into the node features.
Our proposed approach not only overcomes the limitation of the feature-dimension splitting mechanism but also enhances the expressiveness of the transformation by allowing bidirectional information flow between vertex features and learnable queries.

\noindent
\textbf{Relation of Conditional and Unconditional Normalizing Flows.} We observe that the learning objective of our Conditional Graph Normalizing Flow converges to similar optima as the original objective of the unconditional model. Let us consider the evidence lower bound of the unconditional distribution $p(\mathbf{X})$ as $\log p(\mathbf{X}) \geq \mathbb{E}_{q(\mathbf{\bar{X}}|\mathbf{X})}\log p(\mathbf{X}|\mathbf{\bar{X}}) - \mathcal{D}_{KL}(q(\mathbf{\bar{X}}|\mathbf{X}) || p(\mathbf{\bar{X}})$, 
where $\mathcal{D}_{KL}$ is the Kullback–Leibler (KL) divergence.
It should be noted that since $\mathbf{\bar{X}}$ is the learnable queries, it will not be necessary to sample from a distribution $q(\mathbf{\bar{X}}|\mathbf{X})$, i.e., $\mathbb{E}_{q(\mathbf{\bar{X}}|\mathbf{X})}\log p(\mathbf{X}|\mathbf{\bar{X}}) = \log p(\mathbf{X}|\mathbf{\bar{X}})$. In addition, the distribution $q(\mathbf{\bar{X}}|\mathbf{X})$ will be considered a delta function due to the deterministic of $\mathbf{\bar{X}}$, i.e., $q(\mathbf{\bar{X}}|\mathbf{X}) = \delta(\mathbf{\bar{X}} - \mathbf{\bar{X}}_{\theta})$ where $\mathbf{\bar{X}}_{\theta}$ is the optimal point of the learnable queries.
The delta function only contributes at the point $\mathbf{\bar{X}} = \mathbf{\bar{X}}_{\theta}$. 
Therefore, the KL divergence can be rewritten as in Eqn. \eqref{eq:kldiv}.
\begin{equation} \label{eq:kldiv}
\begin{split}
    \mathcal{D}_{KL}(q(\mathbf{\bar{X}}|\mathbf{X}) || p(\mathbf{\bar{X}})) = \int \delta(\mathbf{\bar{X}}-\mathbf{\bar{X}}_{\theta})\log \frac{\delta(\mathbf{\bar{X}}-\mathbf{\bar{X}}_{\theta})}{p(\mathbf{\bar{X}})}d\mathbf{\bar{X}} 
 = \log \frac{1}{p(\mathbf{\bar{X})}} = \operatorname{constant}
\end{split}
\end{equation}
As a result, the negative log-likelihood objective of the conditional model yields gradient updates that are similar to those of the unconditional model, as the learning objective of the conditional normalizing flow serves as a tight upper bound on that of the unconditional model.
\begin{equation}
\begin{split}
    - \log p(\mathbf{X}) &\leq -\log p(\mathbf{X}|\mathbf{\bar{X}}) + \operatorname{constant} \quad
    \Rightarrow -\log p(\mathbf{X})  = \operatorname{O}(-\log p(\mathbf{X}|\mathbf{\bar{X}}))
\end{split}
\end{equation}
where $\operatorname{O}$ is the Big-O notation.
Since the negative log-likelihood objective of the conditional model leads to gradient updates that are closely aligned with those of the unconditional model, the conditional model naturally inherits the training stability and convergence behavior of its unconditional one.
In addition, the close relationship between the two learning objectives ensures that the conditional model is optimizing a similarly structured loss landscape, reducing the risk of instability or divergence during training. As a result, our conditional normalizing flow enables a smooth and reliable extension from the unconditional setting, improving both the model's effectiveness and its theoretical guarantees.

\section{Experiments}

\begin{table}[b!]
\centering
\vspace{-4mm}
\caption{\textbf{Comparisons with Prior SoTA Pre-training Methods (Accuracy).}}
\setlength{\tabcolsep}{10pt}
\label{tab:finetune}
\vspace{-3mm}
    \small
    \centering
    \resizebox{1.0\linewidth}{!}{
        \centering
        \begin{tabular}{ l | c  c c c  c c c  c c c  c }
            \hline
            {Method} & \rotatebox{90}{Citeseer} & \rotatebox{90}{Cora} & \rotatebox{90}{Pubmed} & \rotatebox{90}{Cora-CA} & \rotatebox{90}{DBLP-P} & \rotatebox{90}{DBLP-A} & \rotatebox{90}{AMiner} & \rotatebox{90}{IMDB} & \rotatebox{90}{MN-40} & \rotatebox{90}{20News} & \rotatebox{90}{House} \\
            \hline
            \multicolumn{12}{c}{Fine-Tuning Protocol - Semi-Supervised and Supervised Pre-training} \\
            \hline
                        
            MLP \cite{feng2019hypergraph} & 32.5  & 27.9  & 62.1  & 34.8  & 73.5  & 56.0  & 22.3  & 39.1  & 89.4  & 73.1  &  72.2   \\
            
            HGNN \cite{feng2019hypergraph} & 41.9  & 50.0  & 72.9  & 50.2  & 85.3  & 67.1  & 30.3  & 42.2  & 88.0  & 76.4  & 52.7   \\
            
            HyperGCN \cite{yadati2019hypergcn} & 31.4  & 33.1  & 63.5  & 37.1  & 53.5  & 68.2  & 26.4  & 37.9  & 55.1  & 67.0  & 49.8   \\
            
            HNHN \cite{dong2020hnhn} & 43.1  & 50.0  & 72.1  & 48.3  & 84.6  & 62.6  & 30.0  & 42.3  & 86.1  & 74.2  & 49.7   \\
            
            UniGCN \cite{huang2021unignn} & 44.2  & 49.1  & 74.4  & 51.3  & 86.9  & 65.1  & 32.7  & 41.6  & 89.1  & 77.2  & 51.1   \\
            
            UniGIN \cite{huang2021unignn} & 40.4  & 47.8  & 69.8  & 48.3  & 83.4  & 63.4  & 30.2  & 41.4  & 88.2  & 70.6  & 51.1   \\
            
            UniGCNII \cite{huang2021unignn} & 44.2  & 48.5  & 74.1  & 54.8  & 87.4  & 65.8  & 32.5  & 42.5  & 90.8  & 70.9  & 50.8   \\
            
            AllSet \cite{chien2021you}& 43.5  & 47.6  & 72.4  & 57.5  & 85.9  & 65.3  & 29.3  & 42.3  &  92.1  & 71.9  & 54.1   \\
            
            ED-HNN \cite{wang2022equivariant} & 40.3  & 47.6  & 72.7  & 54.8  & 86.2  & 65.8  & 30.0  & 41.4  & 90.7  & 76.2  &  71.3   \\
            
            PhenomNN \cite{wang2023hypergraph} & 49.8  & 56.4  & 76.1  & 60.8  & 88.1   & 72.3  & 33.8   & 44.1  &  95.9  & 74.0  & 70.4   \\
            
            \hline
            \multicolumn{12}{c}{Fine-Tuning Protocol - Unsupervised Pre-training} \\
            \hline

        GraphMAE2 \cite{hou2023graphmae2} & 41.1 & 49.3  & 72.9  & 55.4  & 86.6  & 69.5  & 32.8  & 43.3  & 90.1  & 71.9  & 52.8   \\

MaskGAE \cite{li2023s}& 49.6  & 57.1  & 72.8  & 57.8  & 86.3  & 74.8  & 33.7  & 44.5  & 90.0  & $-$ & 51.8   \\

TriCL \cite{lee2023m} & 51.7   & 60.2   & 76.2  & 64.3   & 88.0  & 79.7  & 33.1  &  46.9  & 90.3  & 77.2  & 69.7   \\
   
HyperGCL \cite{wei2022augmentations} & 47.0  & 60.3  & 76.8   & 62.0  & 87.6  & 79.7   & 33.2  & 43.9  & 91.2  &  77.8  & 69.2   \\
   
H-GD \cite{zheng2022rethinking} & 45.4  & 50.6  & 74.5  & 58.8  & 87.3  & 75.1  & 32.6  & 43.0  & 90.0  &  77.2  & 69.7 \\

HyperGRL \cite{du2022self} & 42.3  & 49.1  & 73.0  & 55.8  & 86.7  & 70.8  & 33.0  & 43.1  & 90.1  & $-$ & 52.5  \\
HypeBoy \cite{kim2024hypeboy} & 56.7   & 62.3   & 77.0   & 66.3   & 88.2   & 80.6   & 34.1   & 47.6   & 90.4  &  77.6  &  70.4    \\

 Graph-VAVE \cite{kipf2016variational}   & 55.0 & 59.6 & 73.4 & 62.9 & 86.8 & 79.5 & 33.4 & 46.5 & 86.5 & 77.0 & 67.5 \\
 Graph-RNN \cite{you2018graphrnn}    & 57.6 & 62.4 & 73.9 & 63.1 & 87.7 & 80.6 & 33.6 & 47.8 & 89.9 & 77.3 & 68.1 \\
 GRevNets-GNF \cite{liu2019gnf} & 57.4 & 63.2 & 74.5 & 66.1 & 88.4 & 81.9 & 33.8 & 49.0 & 91.1 & 79.3 & 69.4 \\
 \hline
 
 {Uncond $\alpha$-Graph}        & 64.0 & 66.4 & 81.0 & 70.0 & 89.2 & 86.5 & 37.5 & 51.7 & 91.8 & 79.4 & 74.2 \\
 
 \textbf{Cond $\alpha$-Graph}          & \textbf{65.9} & \textbf{67.9} & \textbf{82.7} & \textbf{71.9} & \textbf{91.7} & \textbf{87.9} & \textbf{39.9} & \textbf{53.2} & \textbf{93.0} & \textbf{82.1} & \textbf{76.5} \\
            
            \hline
            \multicolumn{12}{c}{Linear Probing Protocol} \\
            \hline

            GraphMAE2 \cite{hou2023graphmae2} & 29.2  & 37.5  & 55.5  & 38.2  & 75.6  & 57.5  & 27.3  & 36.6  & 89.1  & 62.3  & 51.7   \\
        
        MaskGAE \cite{li2023s} & 47.2   & 56.8  & 62.6  & 56.0  & 84.8  & 75.1  & 33.2  & 44.1  &  90.5  & $-$ & 50.0   \\
        
        TriCL \cite{lee2023m} &  53.3  &  62.1  &  74.5  &  63.6  &  87.1  &  80.9  &  35.0  &  48.0  & 80.0  & 67.2  & 69.1    \\
        
        HyperGCL \cite{wei2022augmentations} & 42.6  & 61.8  & 67.6  & 58.1  & 56.6  & 79.8  & 33.3  & 47.5  & 84.1  &  71.2  & 67.1   \\
        
        H-GD \cite{zheng2022rethinking} & 35.6  & 37.6  & 58.0  & 48.6  & 73.3  & 74.0  & 33.8  & 35.2  & 76.6  &  54.8  & 68.3   \\
           
        HyperGRL \cite{du2022self} & 35.3  & 35.4  & 50.2  & 39.4  & 78.7  & 62.7  & 28.0  & 34.8  &  89.4  & $-$ & 52.0   \\

        HypeBoy \cite{kim2024hypeboy} &  59.6  &  63.5  &  75.0  &  66.0  &  87.9  &  81.2  &  34.3  &  48.8  &  89.2  &  75.7  &  69.4    \\
   
        Graph-VAVE \cite{kipf2016variational}   & 57.6 & 58.5 & 73.7 & 64.5 & 85.0 & 80.7 & 32.9 & 45.9 & 86.1 & 75.5 & 66.2 \\
        Graph-RNN  \cite{you2018graphrnn}  & 58.0 & 59.1 & 74.2 & 65.9 & 85.1 & 80.9 & 33.0 & 47.0 & 88.1 & 76.9 & 66.4 \\
        GRevNets-GNF \cite{liu2019gnf} & 58.3 & 60.1 & 75.4 & 65.7 & 85.4 & 81.4 & 33.5 & 48.1 & 90.0 & 77.4 & 67.9 \\
        
            \hline
            {Uncond $\alpha$-Graph}  & 62.2 & 65.2 & 78.8 & 68.7 & 89.8 & 84.5 & 36.6 & 51.7 & 91.6 & 77.5 & 71.6 \\
            \textbf{Cond $\alpha$-Graph} & \textbf{64.7} & \textbf{68.8} & \textbf{80.5} & \textbf{70.8} & \textbf{91.0} & \textbf{86.9} & \textbf{38.6} & \textbf{52.9} & \textbf{92.9} & \textbf{79.4} & \textbf{74.7} \\
            \hline
        \end{tabular}
        }
\end{table}

\subsection{Datasets and Implementation}

\noindent
\textbf{Benchmarks.} We evaluate our models on the node classification task with two protocols, i.e., \textbf{Fine-Tuning}, \textbf{Linear Probing}, and \textbf{Scene Graph Generation}. 
For Fine-tuning and Linear Probing protocols, we use 11 benchmark graph datasets across different domains of expressing co-citation, co-authorship, computer graphics, movie-actor, news, and political membership relations, including Citeseer \cite{giles1998citeseer}, Cora \cite{mccallum2000automating}, Pubmed \cite{sen2008collective},  DBLP-P \cite{tang2008arnetminer},  DBLP-A \cite{tang2008arnetminer}, AMiner \cite{tang2008arnetminer}, IMDB \cite{yanardag2015deep}, Mondelnet-40 (MN-40) \cite{wu20153d}, 20Newsgroups (20News) \cite{Newsgroups20}, and House \cite{chien2021you}. 
For Scene Graph Generation, we evaluate our model on the Panoptic Scene Graph Generation benchmark \cite{yang2022panoptic}.

\noindent
\textbf{Implementation.}
Our graph normalizing flow network $G$ is composed of $L = 12$ (cross-)invertible attention blocks.
Each (cross-)invertible attention block consists of four proposed attention layers with ActNorm activation \cite{kingma2018glow}, followed by a coupling layer.
To ensure fair comparisons, we adopt a linear layer for $\operatorname{TaskHead}$ for the node classification task and follow the training protocol and hyper-parameter settings from \cite{kim2024hypeboy, hou2023graphmae2}.
All models are trained using the Adam optimizer with a fixed weight decay of $10^{-6}$ and a learning rate of $10^{-3}$. 
For Fine-tuning and Linear Probing downstream tasks, each model is trained for 200 epochs.
For the conditional model, we use learnable query features for downstream tasks. 
The scene graph generation experiments are conducted on 12 NVIDIA L40S. Other experiments are conducted on a single NVIDIA A100.
For Scene Graph Generation, we adopt the implementation of \cite{yang2022panoptic} and use our proposed $\alpha$-Graph as the relation head.

\subsection{Main Results}

\noindent
\textbf{Fine-tuning.} 
Table~\ref{tab:finetune} summarizes the results obtained from end-to-end fine-tuning. Building upon our normalizing flows, the model achieves substantial improvements in classification accuracy across all 11 benchmarks. Compared to both supervised and unsupervised pre-training methods, our approach consistently yields better performance.
In particular, our unconditional normalizing flow achieves an average accuracy of $72.0\%$. The performance improves by an additional $3.9\%$ with our proposed conditional model with learnable queries.
These results have confirmed that our approach captures graph-structured representations and serves as an effective initialization for downstream tasks.

\noindent
\textbf{Linear Probing.}
Table~\ref{tab:finetune} presents the performance of our model under the linear probing evaluation protocol. Compared to prior SoTA methods, our proposed approaches consistently improve classification accuracy across all 11 benchmarks. These results highlight the strong representational capacity of our proposed graph normalizing flows, even when the encoder parameters are kept fixed. 
Notably, our conditional normalizing flows with learnable queries outperform existing graph modeling baselines under linear evaluation, achieving an average accuracy of $72.9\%$. These results demonstrate the effectiveness of our proposed approach in capturing relational structures in graph data.

\begin{wraptable}{r}{0.5\linewidth}
 \centering
 \vspace{-4mm}
 \caption{
 \textbf{Comparisons with Prior SoTA Methods on Scene Graph Generation (Recall Metric).}
 }
 \label{tab:sota-psg}
 \resizebox{\linewidth}{!}{%
 \begin{tabular}{l|cccc}
 \hline
 {Method} &
 \multicolumn{1}{c}{{R/mR@20}} &
 \multicolumn{1}{c}{{R/mR@50}} &
 \multicolumn{1}{c}{{R/mR@100}} \\ 
 \hline
 \multirow{1}{*}{{IMP~\cite{xu2017scene}}} 
 & 16.5 / 6.52 & 18.2 / 7.05 & 18.6 / 7.23 \\ 
 \multirow{1}{*}{{MOTIFS~\cite{zellers2018neural}}}  
  &  20.0 / 9.10 & 21.7 / 9.57 & 22.0 / 9.69 \\ 
 \multirow{1}{*}{{VCTree~\cite{tang2019learning}}} 
  &  {20.6} / {9.70} & 22.1 / 10.2 & 22.5 / 10.2 \\ 
 \multirow{1}{*}{{GPSNet~\cite{lin2020gps}}} 
  & 17.8 / 7.03 & 19.6 / 7.49 & 20.1 / 7.67 \\ 
\multirow{1}{*}{{PSGFormer~\cite{yang2022panoptic}}} & 18.6 / 16.7 & 20.4 / {19.3} & 20.7 / {19.7}\\ 
 \multirow{1}{*}{{HIG}}~\cite{nguyen2024hig} 
  & 19.4 / 6.42 & {22.3} / 8.13  & {26.3} / 9.70  \\ 
  \hline
  \textbf{Uncond $\alpha$-Graph} & 
  23.3 / 9.56 & 26.1 / 13.00 & 27.3 / 13.40 \\
  \textbf{Cond $\alpha$-Graph} &  
  \textbf{24.9/ 11.46} & \textbf{27.8 / 15.73} & \textbf{29.0 / 16.16} \\
  \hline
 \end{tabular}%
}
\end{wraptable}

\textbf{Scene Graph Generation.}
Table \ref{tab:sota-psg} compares our proposed $\alpha$-Graph models with prior methods on the PSG benchmark. As shown, both the unconditional and conditional variants of $\alpha$-Graph achieve consistent improvements across all evaluation metrics (R/mR@20, @50, @100). In particular, the unconditional $\alpha$-Graph gains 
23.3 / 9.56, 26.1 / 13.00, and 27.3 / 13.40 
at  R/mR@20, 50, 100, respectively, outperforming prior graph-based methods, \eg HIG and PSGFormer, by a notable margin. The conditional $\alpha$-Graph further boosts performance to 
24.9 / 11.46, 27.8 / 15.73, and 29.0 / 16.16, 
demonstrating the benefits of incorporating learnable queries via cross-attention conditioning. These results confirm that our invertible attention-based formulation effectively enhances relational reasoning.

\subsection{Ablation Studies}

In this section, 
we will analyze the effectiveness of our approach using the linear probing protocol.

\begin{wraptable}{r}{0.6\linewidth}
\vspace{-4mm}
    \caption{\textbf{Effectiveness of Coupling Layers.}} \label{tab:coupling-layers}
    \small
    \setlength{\tabcolsep}{2pt}
    \centering
    \vspace{-2mm}
    \resizebox{1.0\linewidth}{!}{
    \renewcommand{\arraystretch}{1.0}
        \centering
        \begin{tabular}{ l | c  c c c  c c c  c c c  c }
        \hline
         {Method} & \rotatebox{90}{Citeseer} & \rotatebox{90}{Cora} & \rotatebox{90}{Pubmed} & \rotatebox{90}{Cora-CA} & \rotatebox{90}{DBLP-P} & \rotatebox{90}{DBLP-A} & \rotatebox{90}{AMiner} & \rotatebox{90}{IMDB} & \rotatebox{90}{MN-40} & \rotatebox{90}{20News} & \rotatebox{90}{House} \\
            \hline
            Affine Coupling \cite{dinh2016density}     & 54.1 & 55.8 & 69.1 & 61.4 & 75.7 & 75.7 & 28.2 & 38.1 & 71.6 & 66.8 & 53.5 \\
Glow \cite{kingma2018glow}        & 56.6 & 58.6 & 69.9 & 61.9 & 76.1 & 76.1 & 29.2 & 39.6 & 76.5 & 68.6 & 57.5 \\
AttnFlow \cite{sukthanker2022generative}    & 57.2 & 59.3 & 73.4 & 63.9 & 80.7 & 78.5 & 30.0 & 45.8 & 80.7 & 73.7 & 62.4 \\
Flow++ \cite{ho2019flow++}       & 58.9 & 62.8 & 74.4 & 64.9 & 84.7 & 79.2 & 31.1 & 48.3 & 85.3 & 74.2 & 63.4 \\
\hline
Invertible Attention      & 62.2 & 65.2 & 78.8 & 70.8 & 89.8 & 84.5 & 36.6 & 51.7 & 91.6 & 77.5 & 71.6 \\
\textbf{Invertible Cross-Attention} & \textbf{64.7} & \textbf{68.8} & \textbf{80.5} & \textbf{70.8} & \textbf{91.0} & \textbf{86.9} & \textbf{38.6} & \textbf{52.9} & \textbf{92.9} & \textbf{79.4} & \textbf{74.7} \\
        
            \hline

        \end{tabular}
        }
\end{wraptable}

\textbf{Effectiveness of Different Coupling Layers.}
To evaluate the effectiveness of the proposed invertible (cross-)attention layers, we perform ablative experiments on current flow-based models, including the Affine Coupling Layer \cite{dinh2016density}, Glow \cite{kingma2018glow}, Flow++ \cite{ho2019flow++}, and AttnFlow \cite{sukthanker2022generative}. 
The original coupling layers \cite{dinh2016density, kingma2018glow} adopted 2D Convolution in their design. 
To ensure compatibility with graph-structured input, we convert all 2D convolutions into Graph Convolutions. 
As shown in Table~\ref{tab:coupling-layers}, our invertible cross-attention layers consistently outperform prior coupling mechanisms. 
In particular, our approach achieves an average accuracy of $72.9\%$ over 11 benchmarks, demonstrating its strong capacity to capture complex relational structures.

\noindent
\textbf{Effectiveness of Learnable Queries.}
The last two lines in Table~\ref{tab:coupling-layers} report the performance of the unconditional and conditional models with learnable queries.
Using the unconditional model with invertible attention mechanisms yields an average classification accuracy of $70.9\%$ on 11 benchmarks. 
Further improvements are observed when the conditional model is augmented with learnable queries with invertible cross-attention, reaching an average accuracy of $72.9\%$ over 11 datasets. These results demonstrate the effectiveness of the proposed learnable queries with the cross-attention mechanism in capturing graph-structured correlations.

\begin{wraptable}[9]{r}{0.5\linewidth}
\vspace{-4mm}
\caption{\textbf{Effectiveness of Model Size.}} \label{tab:model-size}
    \small
    \centering
    \setlength{\tabcolsep}{2pt}
    \vspace{-2mm}
    \resizebox{1.0\linewidth}{!}{
    \renewcommand{\arraystretch}{1.0}
        \centering
        \begin{tabular}{ c | c  c c c  c c c  c c c  c }
        \hline
         {\# Blocks} & \rotatebox{90}{Citeseer} & \rotatebox{90}{Cora} & \rotatebox{90}{Pubmed} & \rotatebox{90}{Cora-CA} & \rotatebox{90}{DBLP-P} & \rotatebox{90}{DBLP-A} & \rotatebox{90}{AMiner} & \rotatebox{90}{IMDB} & \rotatebox{90}{MN-40} & \rotatebox{90}{20News} & \rotatebox{90}{House} \\

        \hline
        \multicolumn{12}{c}{Conditional Normalizing Flows} \\
        \hline

        6 & 58.9 & 64.2 & 76.7 & 66.2 & 87.7 & 86.6 & 35.4 & 49.9 & 89.6 & 76.6 & 70.5 \\
        8 & 60.8 & 65.3 & 78.8 & 68.8 & 88.8 & 87.6 & 37.0 & 51.4 & 91.4 & 78.9 & 71.8 \\
        12 & 64.7 & 68.8 & 80.5 & 70.8 & 91.0 & 86.9 & 38.6 & 52.9 & 92.9 & 79.4 & 74.7 \\
        \textbf{16} & \textbf{65.6} & \textbf{70.6} & \textbf{81.7} & \textbf{71.8} & \textbf{92.2} & \textbf{88.0} & \textbf{39.2} & \textbf{53.3} & \textbf{93.4} & \textbf{81.2} & \textbf{75.4} \\
            \hline

        \end{tabular}
        }
\vspace{-4mm}
\end{wraptable}

\noindent
\textbf{Effectiveness of Network Size.}
Table~\ref{tab:model-size} presents the performance of our approach with varying numbers ($L$) of invertible (cross-)attention blocks.
We conduct our experiments on the conditional normalizing flows with learnable queries.
Our results indicate that increasing network depth generally improves node classification performance. In particular, with $L = 16$ blocks, the model achieves the highest average accuracy of $73.9\%$ across 11 benchmarks. Although shallow models reduce computational overhead, deeper models capture correlations more effectively.

\begin{wraptable}[12]{r}{0.6\linewidth}
    \vspace{-4mm}
    \caption{\textbf{Effectiveness of Graph Structure.}} \label{tab:graph-structure}
    \vspace{-2mm}
    \small
    \centering
    \setlength{\tabcolsep}{2pt}
    \resizebox{1.0\linewidth}{!}{
    \renewcommand{\arraystretch}{1.0}
        \centering
        \begin{tabular}{ l | c  c c c  c c c  c c c  c }
        \hline
         {Method} & \rotatebox{90}{Citeseer} & \rotatebox{90}{Cora} & \rotatebox{90}{Pubmed} & \rotatebox{90}{Cora-CA} & \rotatebox{90}{DBLP-P} & \rotatebox{90}{DBLP-A} & \rotatebox{90}{AMiner} & \rotatebox{90}{IMDB} & \rotatebox{90}{MN-40} & \rotatebox{90}{20News} & \rotatebox{90}{House} \\
            \hline

        \hline
        \multicolumn{12}{c}{Unconditional Normalizing Flows}\\
        \hline
            
        Full Attention & 56.3 & 58.7 & 74.0 & 61.8 & 84.4 & 76.2 & 30.3 & 46.0 & 86.2 & 69.9 & 63.1 \\
        
        With $\mathbf{W}_{\operatorname{low}}$ & 59.8 & 63.4 & 75.3 & 65.9 & 87.7 & 82.8 & 34.4 & 49.1 & 88.6 & 76.1 & 69.0 \\

        With $\mathbf{W}_{\operatorname{low}}$+$\mathbf{W}_{\operatorname{up}}$ & 
        \textbf{62.2} & \textbf{65.2} & \textbf{78.8} & \textbf{68.7} & \textbf{89.8} & \textbf{84.5} & \textbf{36.6} & \textbf{51.7} & \textbf{91.6} & \textbf{77.5} & \textbf{71.6} \\

        \hline
        \multicolumn{12}{c}{Conditional Normalizing Flows} \\
        \hline

        Full Attention & 58.1 & 60.0 & 76.2 & 64.0 & 83.3 & 83.8 & 32.2 & 46.6 & 86.3 & 71.1 & 62.6 \\

        With $\mathbf{W}_{\operatorname{low}}$ & 62.0 & 65.6 & 79.0 & 68.7 & 88.0 & 84.7 & 36.0 & 50.3 & 89.8 & 77.6 & 71.7 \\

        With $\mathbf{W}_{\operatorname{low}}$+$\mathbf{W}_{\operatorname{up}}$ & \textbf{64.7} & \textbf{68.8} & \textbf{80.5} & \textbf{70.8} & \textbf{91.0} & \textbf{86.9} & \textbf{38.6} & \textbf{52.9} & \textbf{92.9} & \textbf{79.4} & \textbf{74.7} \\
        
        \hline

        \end{tabular}
        }
\end{wraptable}

\noindent
\textbf{Effectiveness of Graph Structure.}
Table~\ref{tab:graph-structure} compares model variants with increasing levels of graph-structure awareness: (1) Full Attention (no structural prior), (2) with $\mathbf{W}_{\operatorname{low}}$ only, and (3) with both $\mathbf{W}_{\operatorname{low}}$ and $\mathbf{W}_{\operatorname{up}}$. To ensure invertibility in the Full Attention setting, we apply an upper triangular identity mask to the attention matrix.
For our conditional $\alpha$-Graph with learnable queries, the model without structural priors only achieves $65.8\%$ accuracy across 11 benchmarks. Introducing partial structural guidance via $\mathbf{W}_{\operatorname{low}}$ raises performance to $70.3\%$, while full graph awareness yields the highest accuracy of $72.9\%$. Similar improvements are also observed in the unconditional setting, highlighting the important role of graph structure.

\begin{wraptable}[4]{r}{0.6\linewidth}
\vspace{-4mm}
\caption{\textbf{Graph Prediction Task (MMD Metric).}} \label{tab:graph-gen}
\vspace{-2mm}
   \small
    \centering
    \resizebox{\linewidth}{!}{
    \renewcommand{\arraystretch}{1.0}
        \centering
        \begin{tabular}{ l | c c c | c c c }
        \hline
         & \multicolumn{3}{c|}{\textbf{Com Small}} & \multicolumn{3}{c}{\textbf{Ego Small}} \\
         \cline{2-7}
         & Degree & Cluster & Orbit & Degree & Cluster & Orbit \\
        \hline
        \hline
            
        GraphVAE \cite{simonovsky2018graphvae} & 0.350 & 0.980 & 0.540 & 0.130 & 0.170 & 0.050 \\
        
        GraphRNN \cite{you2018graphrnn} & 0.080 & 0.120 & 0.040 & 0.090 & 0.220 & 0.003 \\
        
        GRevNets-GNF \cite{liu2019gnf} & 0.200 & 0.200 & 0.110 & 0.030 & 0.100 & \textbf{0.001} \\
        
        \textbf{\(\alpha\)-Graph} & \textbf{0.105} & \textbf{0.150} & \textbf{0.094} & \textbf{0.220} & \textbf{0.055} & \textbf{0.001} \\
        
        \hline
        \end{tabular}
        }
      \vspace{-6mm}
\end{wraptable}

\noindent
\textbf{Performance on Other Graph Learning Tasks.}
Table~\ref{tab:graph-gen} reports degree/cluster/orbit statistics on \textsc{Community Small} and \textsc{Ego Small} using the maximum mean discrepancy (MMD) metric.
As shown in the results, our approach achieves a competitive performance on the graph generation task compared to prior methods.
Table~\ref{tab:edge-prediction} summarizes the accuracy results of Edge Prediction on Cora, Citeseer, and Pubmed.
$\alpha$-Graph achieves \textbf{97.5/98.3/99.3}\% accuracy, exceeding GCN (95.0/95.9/98.7), GAT (93.9/96.3/98.2), GraphSAGE (95.6/97.4/98.9), and GAE (95.1/97.1/97.5).
This corresponds to margins of +1.9, +0.9, and +0.4 over the prior baseline (GraphSAGE).
In addition, we evaluate our model on the graph generation task. Following the GRevNet protocol, we generate with full attention (structure unknown during generation) and predict the final graph via an auto-encoder, mirroring the GRevNet pipeline.

\begin{wraptable}[8]{r}{0.6\linewidth}
\vspace{-4mm}
\caption{\textbf{Edge Prediction Task (Accuracy).}} \label{tab:edge-prediction}
    \small
    \vspace{-2mm}
    \centering
    \resizebox{1.0\linewidth}{!}{
        \centering
        \begin{tabular}{ l | c  c c c  c c c  c c c  c }
        \hline
         {Method} & {{Cora}} & {{Citeseer}} & {{Pubmed}} \\
            \hline

        \hline
            
        GCN \cite{kipf2016semi} & 95.0  & 95.9 & 98.7 \\
        
        GAT \cite{velivckovic2017graph} & 93.9 & 96.3 & 98.2   \\

        Graph-SAGE \cite{hamilton2017inductive} & 95.6 & 97.4 & 98.9 \\
         GAE \cite{salehi2019graph}& 95.1 & 97.1 &97.5 \\
       \textbf{\(\alpha\)-Graph} & \textbf{97.5} & \textbf{98.3} & \textbf{99.3} \\

        \hline

        \end{tabular}
        }
\end{wraptable}

\noindent
\textbf{Effectiveness on Larger-scale Graph.}
To evaluate scalability beyond standard-scale datasets, we conduct additional experiments of the OGBN-Arxiv benchmark \cite{hu2020open}, which contains $169$K nodes. Table~\ref{tab:large-graph} shows that \(\alpha\)-Graph attains \textbf{73.06}\% accuracy, surpassing MLP (55.50\%), Node2Vec (70.07\%), GCN (71.74\%), GAT (71.59\%), GraphMAE2 (71.21\%), and GraphSAGE (71.49\%).The margin over the strongest competing baseline (GCN) is \textbf{+1.32} points, gains relative to other baselines are +2.99 (Node2Vec), +1.47 (GAT), +1.85 (GraphMAE2), and +1.57 (GraphSAGE), indicating that our approach remains effective at this scale and is not confined to standard-size graphs.

\begin{table}
\caption{\textbf{Effectiveness on Larger-scale Graph.}} \label{tab:large-graph}
    \small
    \centering
    \setlength{\tabcolsep}{2pt}
    \resizebox{1.0\linewidth}{!}{
    \renewcommand{\arraystretch}{1.0}
        \centering
        \begin{tabular}{  c  c c c  c c c  c c c  c }
        \hline
         \rotatebox{0}{{{MLP}}} & 
         \begin{tabular}{@{}c@{}}Node2Vec \\ \cite{grover2016node2vec} \end{tabular} & 
         \begin{tabular}{@{}c@{}}GCN \\ \cite{kipf2016semi} \end{tabular} & 
         \begin{tabular}{@{}c@{}}GAT \\ \cite{velivckovic2017graph} \end{tabular} & 
         \begin{tabular}{@{}c@{}}G-MAE2 \\ \cite{hou2023graphmae2} \end{tabular} & 
         \begin{tabular}{@{}c@{}}G-SAGE \\ \cite{hamilton2017inductive} \end{tabular} & 
         \rotatebox{0}{\textbf{{\(\alpha\)-Graph}}}  \\
            \hline

        55.50 & 70.07 & 71.74 & 71.59 & 71.21 & 71.49 & \textbf{73.06}  \\

        \hline

        \end{tabular}
        }
        \vspace{-6mm}
\end{table}

\begin{wraptable}{r}{0.5\linewidth}
\caption{\textbf{Effectiveness of a learnable \(\boldsymbol{d}\).}} \label{tab:learnable-d}
    \small
    \centering
    \resizebox{1.0\linewidth}{!}{
    \renewcommand{\arraystretch}{1.0}
        \centering
        \begin{tabular}{ l | c  c c c  c c c  c c c  c }
        \hline
          & {{Citeseer}} & {{Cora}} & {{Pubmed}} & {{Cora-CA} }  \\
            \hline

        Static & 63.4\%  & 67.5\% & 78.8\% & 68.0\% \\
        \hline
       \textbf{Learnable} & \textbf{64.7\%} & \textbf{68.8\%} & \textbf{80.5\%} & \textbf{70.8\%}\\

        \hline

        \end{tabular}
        }
\end{wraptable}

\noindent
\textbf{Effectiveness of a Learnable Variable \(\boldsymbol{d}\).}
We ablate the attention parameter \(d\) by comparing a fixed (static) value with a learnable variant.
Table~\ref{tab:learnable-d} shows consistent gains across four datasets: Citeseer \(64.7\%\) vs. \(63.4\%\) \((+\!1.3)\), Cora \(68.8\%\) vs. \(67.5\%\) \((+\!1.3)\), Pubmed \(80.5\%\) vs. \(78.8\%\) \((+\!1.7)\), and Cora-CA \(70.8\%\) vs. \(68.0\%\) \((+\!2.8)\).
On average, making \(d\) learnable yields \(+\!1.78\) points (\(+2.56\%\) relative).
This supports the intuition that learning \(d\) allows the model to adapt the sharpness of the attention distribution, improving node-classification performance.

\begin{wraptable}{r}{0.5\linewidth}
\caption{\textbf{Computational Efficiency.}} \label{tab:computatinal}
  \small
    \centering
    \resizebox{1.0\linewidth}{!}{
    \renewcommand{\arraystretch}{1.0}
        \centering
        \begin{tabular}{ l  c  c c c  c c c  c c c  c }
        \hline
         \textbf{Method} & \textbf{{Accuracy}} & \textbf{{Inference Time}}  \\
            \hline

        \hline
            
        GAT & 71.59  & 5.5 mins \\
        
        GraphMAE2 & 71.21 & \textbf{5.1 mins} &  \\

        $\alpha$-Graph & \textbf{73.06} & 5.9 mins \\

        \hline

        \end{tabular}
        }
\end{wraptable}

\noindent
\textbf{Computational Efficiency.}
We benchmark inference on the 169K-node OGBN-Arxiv graph against strong baselines.
Table~\ref{tab:computatinal} shows that $\alpha$-Graph attains the best accuracy (\textbf{73.06}\%) while keeping inference time comparable (5.9\,min vs.\ 5.5\,min for GAT and \textbf{5.1}\,min for GraphMAE2).
Relative to GAT, we gain +1.47 points ($\approx$2.0\%) for an extra 0.4\,min ($\approx$7.3\%); relative to GraphMAE2, +1.85 points ($\approx$2.6\%) for +0.8\,min ($\approx$15.7\%).
This reflects a favorable accuracy–latency trade-off, despite a more expressive and invertible architecture, the method remains practical at this scale.
Further speedups are feasible via sparse attention, block-wise computation, or graph partitioning, which we leave to future work.

\section{Conclusions}

\noindent
\textbf{Conclusions.}
This work has introduced a new explicit approach to graph modeling via Normalizing Flows. Our proposed Invertible Cross-Attention efficiently captures the complex relational structure of graph data and addresses the limitations of prior coupling layers. Then, to improve the expressiveness of our model, we present a new Conditional Normalizing Flow model with Invertible Cross-Attention. Our theoretical analysis also shows that the Conditional Normalizing Flow yields optima similar to those of the unconditional model. Our experimental results show the effectiveness of the proposed approach and outperform prior SoTA graph learning methods.

\noindent
\textbf{Limitations.}
In our experiments, we adopt learning hyperparameters and benchmarks to support our hypothesis. Nevertheless, several limitations exist. This study primarily investigates the effectiveness of our proposed graph normalizing flows with the invertible attention mechanism. Due to computational constraints, our experiments are restricted to the standard scale of the selected benchmarks. However, based on the fundamental theories established in this work, we hypothesize that the proposed approaches can generalize effectively to larger-scale graphs and benchmark settings.

\bibliography{references}
\bibliographystyle{tmlr}

\end{document}